\pdfoutput=1

\documentclass[11pt]{article}

\usepackage{acl}

\usepackage{times}
\usepackage{latexsym}
\usepackage{graphicx}
\usepackage{booktabs}
\usepackage{multirow}
\usepackage{amsmath}
\usepackage{bm}
\usepackage{makecell}
\usepackage{algorithm}
\usepackage{algorithmic}
\usepackage{amsfonts}
\usepackage{amssymb}
\usepackage{subfig}
\usepackage{float}

\usepackage[T1]{fontenc}

\usepackage[utf8]{inputenc}

\usepackage{microtype}

\usepackage{inconsolata}

\usepackage{xspace}

\newcommand{\tool}{XMoE\xspace}

\title{\tool: Sparse Models with Fine-grained and Adaptive  Expert Selection}

\author{Yuanhang Yang$^{1}$ \quad Shiyi Qi$^{1}$ \quad Wenchao Gu$^{2}$ \quad Chaozheng Wang$^{2}$ \\ \textbf{Cuiyun Gao}$^{1}$ \quad \textbf{Zenglin Xu}$^{1}$ \\
$^{1}$Harbin Institute of Technology, Shenzhen, China\\ 
$^{2}$CUHK, Hongkong, China \\
\{ysngkil, syqi12138, guwenchaohk\}@gmail.com \quad czwang23@cse.cuhk.edu.hk \\
 \{gaocuiyun, xuzenglin\}@hit.edu.cn \\
}

\begin{document}
\maketitle
\begin{abstract}
Sparse models, including sparse Mixture-of-Experts (MoE) models, have emerged as an effective approach for scaling Transformer models. However, they often suffer from computational inefficiency since a significant number of parameters are unnecessarily involved in computations via multiplying values by zero or low activation values. To address this issue, we present \tool, a novel MoE designed to enhance both the efficacy and efficiency of sparse MoE models. \tool leverages small experts and a threshold-based router to enable tokens to selectively engage only essential parameters. Our extensive experiments on language modeling and machine translation tasks demonstrate that \tool can enhance model performance while decreasing the computation load at MoE layers by over 50\% without sacrificing performance. Furthermore, we present the versatility of \tool by applying it to dense models, enabling sparse computation during inference. We provide a comprehensive analysis and make our code available at \url{https://github.com/ysngki/XMoE}.

\end{abstract}

\section{Introduction}

Recently, remarkable advancements in large language models have been achieved through scaling up their sizes \citep{Brown2020LanguageMA, Chung2022ScalingIL, Touvron2023Llama2O}. However, this progress has come with a significant increase in training costs, posing a challenge to further scaling. To address this issue, sparse models, such as sparse Mixture-of-Experts (MoE) models, have emerged as an alternative approach.
 These models allow for scaling the model size without a corresponding increase in computational cost \citep{ DBLP:conf/iclr/LepikhinLXCFHKS21, jaszczur2021sparse, DBLP:journals/jmlr/FedusZS22, DBLP:conf/icml/RajbhandariLYZA22, DBLP:journals/corr/abs-2402-01739, DBLP:journals/corr/abs-2401-04088}. 
The key to this ability lies in sparsely activated MoE layers. These layers comprise multiple sub-networks, or ``experts'', typically implemented as Feed-Forward Networks (FFNs) \citep{DBLP:conf/nips/VaswaniSPUJGKP17}. Unlike traditional models that utilize all parameters for each input token, MoE models selectively activate a subset of experts. This approach effectively decouples computational costs from model size, paving the way for more efficient scaling.

\begin{figure}
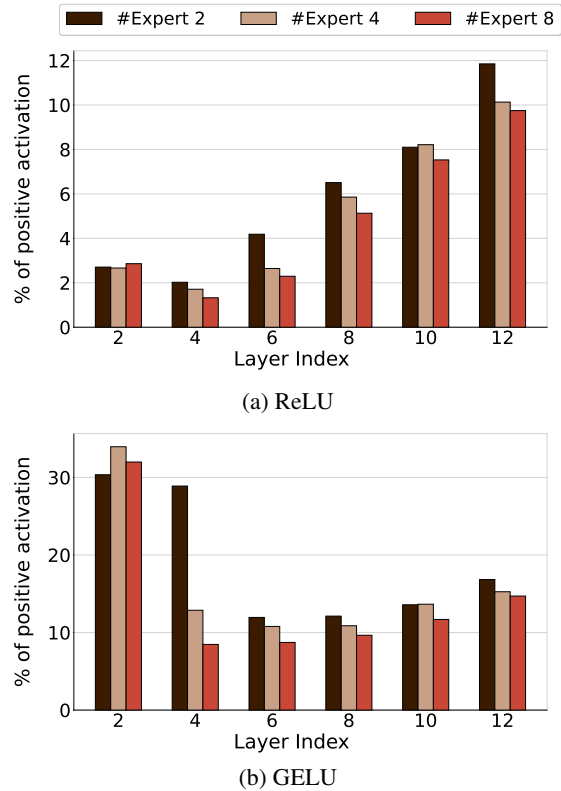

    \centering
    \subfloat[ReLU]{\includegraphics[width=.95\columnwidth]{figures/0_relu_intro_bar_fig.pdf}} \\
    \vspace{-1em}
    \subfloat[GELU]{\includegraphics[width=.95\columnwidth]
    {figures/0_gelu_intro_bar_fig.pdf}}
    \vspace{-0.5em}
    \caption{Average percentage of positive values in the FFN layers after the activation function.
    All models are decoder-only Transformers with 12 layers.}
    \label{fig:0_intro}
\end{figure}

While MoE models are effective for scaling, this paper argues that MoE models exacerbate the issue of computational inefficiencies. Initially identified in the dense model T5 \citep{DBLP:journals/jmlr/RaffelSRLNMZLL20}, this issue is characterized by a significant portion of computations in the FFN layer being wasted on multiplying values by zero  \citep{zhang-etal-2022-moefication, li2023the}. As illustrated in Figure \ref{fig:0_intro}, the computational inefficiency issue is also prevalent in sparse models and even worsens as the number of experts increases. 
This observation suggests that only a small portion of the parameters in an expert is useful for the input, while others are unnecessarily involved in the computation.
Consequently, selecting one expert for each input can already lead to a significant waste of computation. In order to alleviate this problem, a more fine-grained and adaptive strategy for parameter selection is necessary.

\begin{figure}
    \centering
    \includegraphics[width=0.45\textwidth]{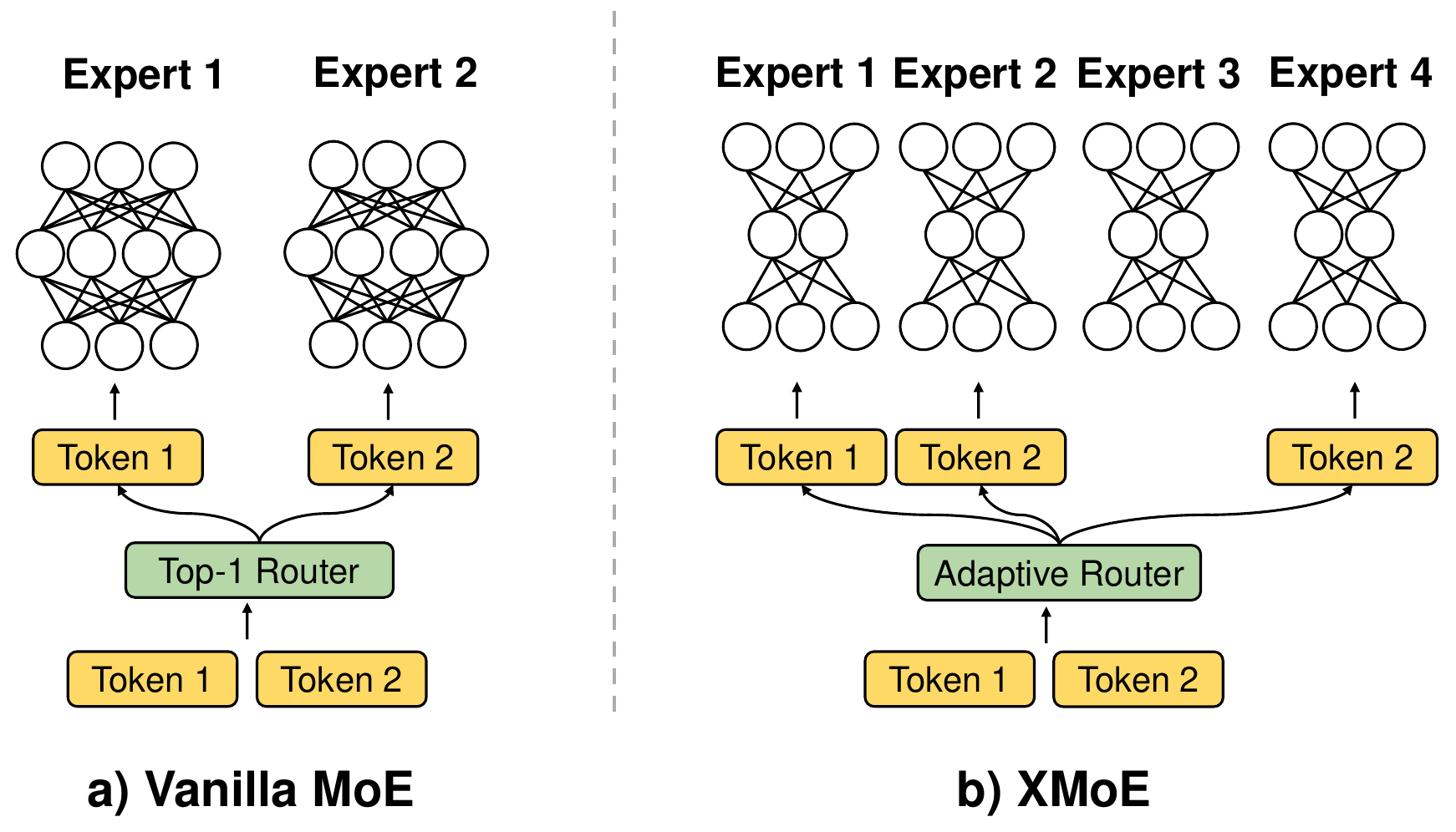}
    \caption{Overview of an MoE layer in \tool, where tokens are routed to small experts by an adaptive router.}
    \label{fig:1_overview}
\end{figure}

Figure  \ref{fig:1_overview} shows the overview of a novel MoE design, named \tool, which allows tokens to select fewer parameters to improve the efficiency without hindering model performance.
To achieve this, \tool proposes to exploit small experts and a threshold-based router. First, considering that expert is the smallest unit of parameter selection in MoE models, utilizing small experts is the prerequisite for the more fine-grained selection. It allows models to choose the useful parameters precisely without activating the redundant parameters. 

In order to ensure the effectiveness, a novel adaptive router is further exploited by \tool.
Different from the widely-used top-$k$ router that dispatches each input to a fixed number of experts, this adaptive router allows tokens to self-determine the number of required experts based on a pre-defined threshold. Intuitively, an easily processed token can be routed to a single small expert, while a critical token may require multiple experts
\citep{DBLP:conf/nips/ZhouLLDHZDCLL22, li-etal-2023-adaptive-gating}.
An adaptive router allows models to leverage the difference of the input complexity to dynamically allocate computational resources.
This not only enhances model efficiency but also yields potential quality improvements when computational resources are constrained.

In conjunction with the aforementioned design, \tool aims to enhance the efficiency of MoE models, with a focus on improving quality with a fixed computational budget or reducing computational costs without compromising performance.
Extensive experiments in language modeling and machine translation demonstrate performance gains over existing MoE methods. \tool also enables a reduction in Floating Point Operations (FLOPs) by over $50\%$ with minimal impact on performance.
Additionally, our investigation extends to training dense models using \tool to facilitate sparse computation during inference. This approach not only matches the performance of dense counterparts but also facilitates a substantial reduction in FLOPs.

Our contributions can be summarized as: (1) We identify a computational inefficiency issue in current sparse MoE models. (2) We propose \tool to improve the efficiency of MoE models with small experts and a novel routing strategy. (3) Extensive experiments in language modeling and machine translation demonstrate \tool as a promising alternative to existing sparse and dense models.

\section{Background}

\subsection{Mixture-of-Experts (MoE)}
MoE is a family of neural network architecture that enables conditional computation by sparsely activating small sub-networks, so-called experts, based on a pre-defined router.
The core component of MoE models is the MoE layer, which consists of a set of experts $\{E_1, \cdots, E_N\}$ and a routing function \cite{DBLP:conf/iclr/ShazeerMMDLHD17, zoph2022stmoe, DBLP:journals/corr/abs-2401-06066}. Each expert $E$ is a parameterized function. The routing function is utilized to route individual tokens to their assigned experts.
In this work, we consider MoE for Transformer, where FFN layers within Transformer are substituted with large MoE layers, in which each expert is an independent FFN. 

\medskip

\noindent \textbf{Trainable Router.}
A commonly used router is based on a gating network consisting of a trainable weight matrix followed by a softmax function.
For the input token $x$ with its intermediate representation denoted as $h \in \mathbb{R}^{d}$, the router computes the probability distribution over experts as:
\begin{align}
    p = \mathrm{Softmax}(W \cdot h) \label{eq:prob},
\end{align}
where $W \in \mathbb{R}^{N \times d}$, $N$ denotes the total number of experts and $d$ denotes the hidden size of the model.
The set of top-$k$ experts $\mathcal{E}$ is decided based on $p$, where $| \mathcal{E} | = k$. 
Each expert processes tokens independently. Then, the final output of the token is the weighted sum of the output of its $k$ experts:
\begin{align}
    y = \sum_{i \in \mathcal{E}} p_iE_i(h). \label{eq:sum}
\end{align}

\medskip

\noindent \textbf{Load Imbalance.}
Learnable routers can easily cause the load imbalance issue, as shown in \citet{DBLP:conf/iclr/ShazeerMMDLHD17}. If there is no constraint employed during training, most tokens would be dispatched to a small number of experts, leading to a large portion of experts being insufficiently trained \citep{DBLP:conf/iclr/LepikhinLXCFHKS21}. In addition, if experts are distributed across different nodes, most nodes must wait for others to finish the computation, thus hindering the training efficiency. 
\medskip

 \noindent \textbf{Capacity.} Expert capacity is introduced to MoE models to avoid the severe load imbalance issue by limiting the maximum number of tokens routed to each expert \citep{DBLP:conf/iclr/LepikhinLXCFHKS21, DBLP:journals/jmlr/FedusZS22, DBLP:conf/icml/RajbhandariLYZA22, puigcerver2024from}. Suppose a given batch's token number is $T$, and the expert number is $N$. The expert capacity $C$ is:
\begin{align}
    C = \frac{T}{N}  \cdot \gamma \label{eq:cap}
\end{align}
where $\gamma$ refers to the preset capacity factor.
If an expert is underutilized, the unused capacity buffers are filled with padding tokens. Once an expert is at capacity, additional tokens are dropped, which means being passed to the subsequent Transformer block directly.
\medskip

\noindent \textbf{MoE for Dense Models.}
Although MoE models are proposed for training large-scale sparse models, recent studies bring the concept of MoE for either pruning or training dense models. \citet{zhang-etal-2022-moefication} propose to divide the feed-forward network layers in a pretrained dense Transformer into several small experts based on heuristic strategies. 
Each input token selectively activates top-$k$ experts instead of utilizing the whole FFNs.
\citet{DBLP:conf/iclr/ChenZJLW23} propose to view MoE as a regularization method for training dense Transformers. 
Although our work proposes \tool for sparse model training, it can also be utilized for training dense models to enable sparse computation during inference.

\subsection{FFN as Memory}
Recent studies suggest that an FFN layer can be conceptualized as a memory layer composed of numerous key-value memory pairs \citep{DBLP:conf/nips/LampleSRDJ19, geva-etal-2021-transformer}. 
In this view, each column in the first matrix of an FFN layer is a key vector, and the corresponding row in the second matrix is the value vector. 
It is observed that only some memory values benefit an input token \citep{dai-etal-2022-knowledge}, leading to most memory pairs contributing to redundant computations. We hypothesize that in MoE models, which incorporate multiple FFNs, the beneficial memories tend to be distributed across different FFNs. 
This dispersion of useful memories diminish overall efficiency.

\section{Method} \label{sec:method}

\subsection{FFN Decomposition}
Considering that widely-used activation functions such as ReLU and GELU operate on an element-wise basis, it is reasonable to conceptualize an FFN layer as a composition of several smaller FFN layers
\begin{align}
    y = W_2\ \sigma(W_1\cdot x) = \sum_{i} W^{i}_2\ \sigma(W^{i}_1\cdot x),
\end{align}
where $W^{i}_j$ is the parameters of $i$-th small FFN layer and $\sigma$ denotes the activation function. Instead of maintaining a single large FFN, \tool trains multiple small FFNs and compose them to produce an output.
In subsequent discussions, we assume that the dimensions are identical for each experts unless stated otherwise.
Our preliminary experiments showed that the GELU activation function performed slightly better. Therefore, we will use GELU by default.

\subsection{Threshold-based Router}
\tool's router consists of a trainable weight matrix $W \in  \mathbb{R}^{d \times N}$,  where each column of $W$ serves as a centroid representing an expert.
Given an input token $x$ and its intermediate representation $h \in \mathbb{R}^d$, the probability of choosing the $i$-th experts is $p_i$, according to the Eq \eqref{eq:prob}.

Considering the distribution $p$ varies with layers, tokens and model configurations, manual selection of an optimal value for the top-$k$ parameter $k$ can be challenging \citep{DBLP:journals/corr/abs-2105-15082, li-etal-2023-adaptive-gating}.
A higher value of $k$ can ensure effectiveness but  at the cost of increased computational overhead per token, potentially impacting efficiency. Conversely, a lower value of $k$ may reduce computational load  but  restrict the model's capacity to handle complex tokens.
To navigate this trade-off, \tool employs a threshold-based selection mechanism, enabling tokens to self-determine the number of experts they should be routed to based on a predefined threshold parameter $t$, which ranges from 0 to 1.

\begin{figure}
    \centering
    \includegraphics[width=0.45\textwidth]{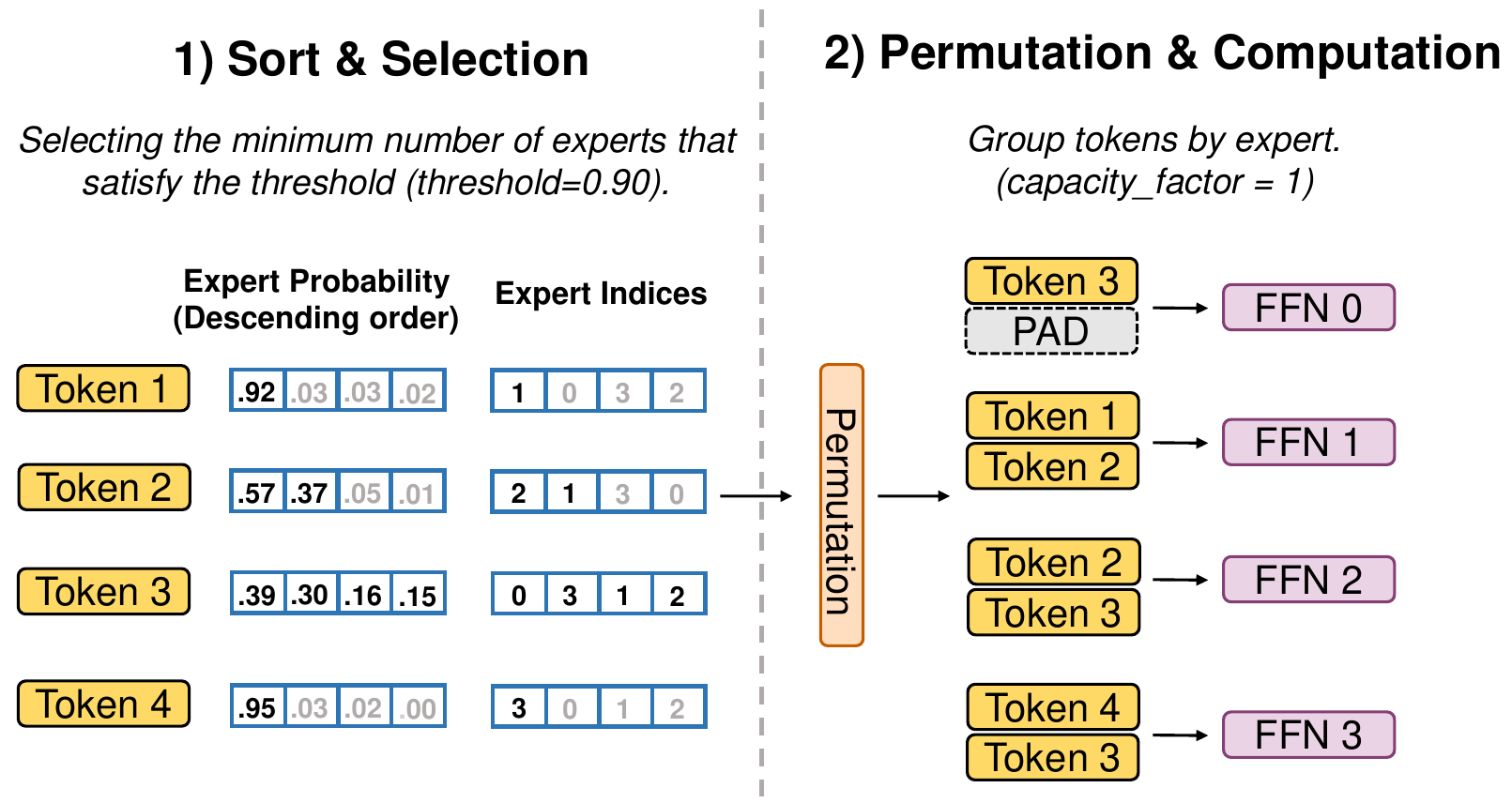}
    \caption{Overview of the threshold-based router. The number of tokens processed per expert is determined according to both the total token number and the capacity factor.}
    \label{fig:2_router}
\end{figure}

The overview of the selection procedure is illustrated in Figure \ref{fig:2_router}. 
Initially, the probabilities $p$ are sorted in descending order: $p = \left[ p_{i_1}, p_{i_2}, ..., p_{i_N} \right]$, where $p_{i_1} \geq p_{i_2} \geq ... \geq p_{i_N}$ and $i_j$ indicates the index of top-$j$ expert. The router then identifies the smallest index $m$ such that the cumulative sum of probabilities up to index $m$ is greater than or equal to the threshold $t$:
\begin{align}
 \underset{m}{\arg\min} \sum^{m}_{j=1} p_{i_j} \geq t.
\end{align}
The input tokens are dispatched to their first $m$ experts. 
According to the Eq \eqref{eq:sum}, an expert's contribution to the output is directly proportional to its assigned probability $p$. Thus a high cumulative sum of probabilities indicates that tokens are routed to the most relevant experts, while other experts with lower probabilities have  minimal influence and can be ignored.
\smallskip

\noindent \textbf{Priority.} 
The threshold-based router in MoE models permits a token to select multiple experts, potentially overwhelming certain experts and leading to dropped tokens due to limited expert capacities.
To mitigate this, \tool assigns a priority $r$ to a token dispatched to a specific expert.
The experts process tokens with higher priorities first.
 Specifically, the prioritization is based on the likelihood of a token considering the expert as its preferred choice, determined by a heuristic rule.
If a token views an expert as its top-$i$ choice with probability $p$, the priority $r$ is set heuristically as $(p - i)$. Appendix \ref{app:exp_sel} presents the algorithmic implementation.

\subsection{Auxiliary Loss}
Following \citet{DBLP:journals/jmlr/FedusZS22}, we utilize an auxiliary loss as a part of the training objective. It encourages input tokens to be uniformly dispatched to the experts. We only impose this constraint on the top-$1$ assignment. The details are provided in Appendix \ref{app:balance_loss}.

\subsection{Complexity}
Within an MoE layer, a total of $C \cdot N$ tokens are processed by the experts, where $C$ denotes the capacity and $N$ is the number of experts. Let $O(E)$ represents the computational complexity associated with an expert handling a single token. The computational complexity of an MoE layer, as indicated by Eq. \ref{eq:cap}, can be expressed as:
\begin{align}
    O(\mathrm{MoE}) \propto  \gamma \cdot T \cdot O(E). \label{eq:complexity}
\end{align}

A small $\gamma$ can reduce the computational cost but aggravate token dropping since experts may not have enough capacity to process received tokens.
On the contrary, increasing $\gamma$  could lead to additional computational overhead and tend to waste computing resources on padding tokens. \tool allows to trade off the efficacy and efficiency by concurrently increasing $\gamma$ and diminishing $O(E)$.

\subsection{Comparison with Top-k Router}
\noindent \textbf{Top-$1$ Router.}
The top-$1$ router exclusively assigns each token to a single expert. It potentially leads to inefficiencies when experts receive fewer tokens than their capacity, resulting in wasted computations processing padding tokens.
In contrast, the threshold-based router dispatch each token to at least one expert. Consequently, the token assignment of the top-$1$ router is a subset of that produced by the threshold-based router. This enables models with the threshold-based router to implicitly reduce computational inefficiencies.

\noindent \textbf{Top-$k$ Router.} Top-$k$ router allows each token to choose multiple experts, thus unlikely leading to computation waste, especially when the capacity is limited.
However, this router processes tokens with the same amount of computation, regardless of the difference in complexity between tokens.
In contrast, an adaptive router allows the model to dynamically allocate computational resources to important tokens while reducing unnecessary computations, leading to better utilization of training computation.

The threshold-based router and the top-$k$ router share parameters of the same shape, with their only distinction lying in their respective selection strategies. 
This structural similarity indicates that the threshold-based routing strategy and the top-$k$ strategy can be seamlessly interchanged without necessitating any modifications.

\subsection{Dense Model Applications}
A dense model can be treated as an MoE model with only one expert per layer. \tool can be applied by decomposing FFN layers in the dense model into multiple smaller ones. In contrast to training a sparse model, we \textit{densely} train this model by setting the threshold $t = 1.0$ and $\gamma$ as $N$. This setup ensures that each token is processed by all experts without token dropping. It is expected that the router can learn to measure the importance of different experts. During inference, both $t$ and $\gamma$ can be adjusted to enable sparse computation.

\section{Experiments}

\subsection{Tasks and Datasets}

\textbf{Language Modeling.} We pretrain models on two English-language datasets, OpenWebText \citep{Gokaslan2019OpenWeb} and Wikitext-103 \citep{DBLP:conf/iclr/MerityX0S17}. The former is a public reproduction of WebText used for GPT-2 training \citep{Radford2019LanguageMA}. The latter is a smaller language modeling corpus containing Wikipedia articles.
\medskip

\noindent \textbf{Machine Translation.} We have collected 5 language pairs from WMT23 datasets. Each language pair is trained independently. The detailed statistics for these language pairs can be found in Table \ref{tab:trans_data}. 

\subsection{Model Configuration}

\noindent \textbf{Language Modeling.} We adopt the Transformer decoder with $12$ layers. Our implementation is based on Megatron-LM \citep{DBLP:journals/corr/abs-1909-08053}. We use the tokenizer of GPT-2 of which the vocabulary size is $50,257$. The model is trained for $60$K steps in total on OpenWebText and $4$k steps on Wikitext-103. The learning rate is $4.5 \times 10^{-4}$ and the cosine learning rate decay is exploited. During training, we use float$16$ for acceleration. The threshold $t$ is set to $0.90$ based on the results of pilot experiments.
\medskip

\noindent \textbf{Machine Translation.} We adopt the Transformer-based architecture with 12 encoder layers and 6 decoder layers. Our implementation is based on fairseq \citep{ott-etal-2019-fairseq}. The vocabulary is learned from the training data for each language pair using byte-pair encoding. The threshold $t$ is set to $0.90$.  Automatic mixed precision is enabled to accelerate training. We report the detokenized BLEU and $\textrm{chrF}2^{++}$ using sacreBLEU\footnote{BLEU:\textit{nrefs:1 | case:mixed | eff:no | tok:13a | smooth:exp | version:2.3.2}. chrF2:\textit{nrefs:1 | case:mixed | eff:yes | nc:6 | nw:2 | space:no | version:2.3.2}}.
\smallskip

For sparse models, an MoE layer is utilized to replace the dense FFN layer in every alternate Transformer block, following previous practice \citep{DBLP:conf/nips/ZhouLLDHZDCLL22} . The capacity factor $\gamma$ is set to $1.0$ for models with top-1 routers. For models with smaller experts, we increase the capacity factor according to Eq \eqref{eq:complexity} to ensure the consumed training compute identical. 
We run our experiments on one node with $4$ NVIDIA A100 GPUs.

\subsection{Baselines}
Top-$1$ routing, which assigns each token to one expert, is widely used in MoE models, such as Switch Transformers \citep{DBLP:journals/jmlr/FedusZS22}, BASE Layers \citep{DBLP:conf/icml/LewisBDGZ21} and Hash Layers \citep{DBLP:conf/nips/RollerSSW21}. Switch Transformer utilizes a learnable top-$1$ router with an auxiliary loss to alleviate the load imbalance issue.
BASE Layer proposes to view the token assignment as a linear assignment problem. 
They force each expert to process an equal number of tokens during training while using greedy assignments at test time.
 Hash Layers replace the learnable router with simple hash functions. We implement Hash Layers with a random hash function, which is suggested to have strong performance \citep{DBLP:conf/nips/RollerSSW21}. We extend the top-$1$ router of Switch Transformer to the top-$k$ ($k > 1$) router.

\section{Results}

\begin{table*}[h!]
\centering
\resizebox{\linewidth}{!}{
\begin{tabular}{clccc|cc}
\toprule[0.10em]
\textbf{Params} &  \textbf{Method} &  \textbf{Top-}\bm{$k$}  & \textbf{\# Experts} & \textbf{Expert Size} & \textbf{OpenWebText}  & \textbf{WikiText-103 }\\
\midrule[0.08em]
$124$M & Dense Transformer & $1$ & $1$ &  $3072$  &  $22.61$ & $22.24$ \\

\midrule[0.08em]


 
 \multirow{9}{*}{$323$M} & Switch Transformer & $1$ & $8$ &  $3072$  &  $20.11$ &  $20.60$ \\
& Hash Layer & $1$ & $8$ &  $3072$     & $21.27$  &  $21.63$ \\
& BASE Layer & $1$ & $8$ &  $3072$    & $20.49$ &  $21.13$ \\
\cmidrule[0.08em](l){2-7} 

& \multirow{3}{*}{Top-$k$ Gating} 
& $2$ & $16$ &  $1536$  & $19.81$ &   $20.24$ \\
& & $4$ & $32$ &  $768$   & $19.75$  &   $20.14$ \\
& & $8$ & $64$ &  $384$     & $19.79$ &  $20.10$ \\
\cmidrule[0.08em](l){2-7} 

& \multirow{3}{*}{\tool} & \multirow{3}{*}{-} & $16$ &  $1536$   & $19.57$ &  $20.16$ \\
& &  & $32$ &  $768$     & $19.49$ &  $20.00$ \\
& &  & $64$ &  $384$     & $\mathbf{19.46}$ &  $\mathbf{19.97}$ \\

\midrule[0.08em]

\multirow{9}{*}{$556$M} & Switch Transformer & $1$ & $8$ &  $4096$  &  $19.17$  &  $19.72$\\
& Hash Layer & $1$ & $8$ &  $4096$    & $20.25$ &  $22.01$\\
& BASE Layer & $1$ & $8$ &  $4096$      & $19.72$ &  $21.03$ \\
\cmidrule[0.08em](l){2-7} 
& \multirow{3}{*}{Top-$k$ Gating} 
& $2$ & $16$ &  $2048$     & $18.94$ & $19.47$ \\
& & $4$ & $32$ &  $1024$     & $18.94$ &   $19.36$ \\
& & $8$ & $64$ &  $512$     & $18.99$ &   $19.15$\\
\cmidrule[0.08em](l){2-7} 
& \multirow{3}{*}{\tool}
& \multirow{3}{*}{-} & $16$ & $2048$    &  $18.72$ &  $19.41$ \\
& &  & $32$ &  $1024$     & $\textbf{18.68}$ &  $19.26$ \\
& & & $64$ &  $512$     & $18.72$ &  $\textbf{19.10}$\\

\bottomrule[0.10em]
\end{tabular}
}
\caption{Test $\textrm{perplexity}^{\downarrow}$ on OpenWebText and WikiText-103. Models consume approximately the same training and inference FLOPs through the adjustment of $\gamma$ according to Eq. \ref{eq:complexity}. The ``Top-$k$'' column denotes the number of selected experts per token, and ``-'' denotes not applicable.}
\label{tab:ppl}
\end{table*}

\subsection{Language Modeling}

Table \ref{tab:ppl} reports the perplexity results of models with equivalent computational complexity in MoE layers.
A consistent performance gain is observed across both the OpenWebText and WikiText-103 datasets when reducing the expert size.
Specifically, when the expert size decreases to $384$, \tool with $323$M parameters achieves the best performance. It surpasses  top-$k$ routing and Switch Transformer by $0.33$ and $0.65$ perplexity points on OpenWebText, respectively. 
The difference between \tool and top-$k$ routing lies in the routing strategy, while the difference between top-$k$ routing and Switch Transformer lies in the expert size.
This suggests that a threshold-based router can assist models in leveraging expert capacities, and smaller experts can enhance model quality.
There are intriguing outcomes when we increases the parameter count to $556$M by expanding the intermediate dimension of experts.
It is shown that top-$k$ routing with a size of 1024 outperforms that with a size of 512 on OpenWebText, as does \tool. We attribute this to the optimization of sparse models and leave this for future investigation.

\begin{figure}[h!]
    \centering
    \subfloat[OpenWebText]{\includegraphics[width=.90\columnwidth]{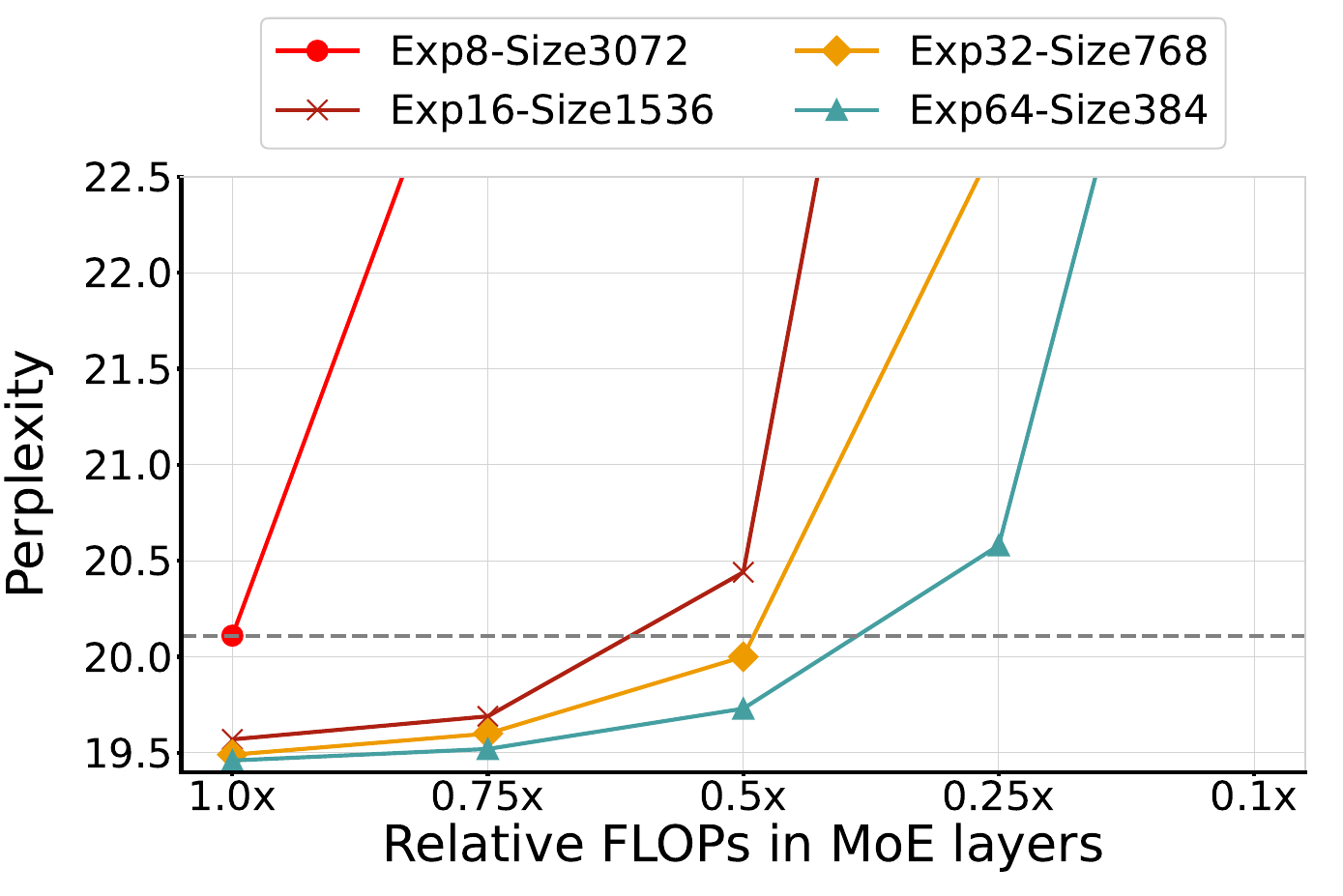}} \\
    \subfloat[WikiText-103]{\includegraphics[width=.90\columnwidth]
    {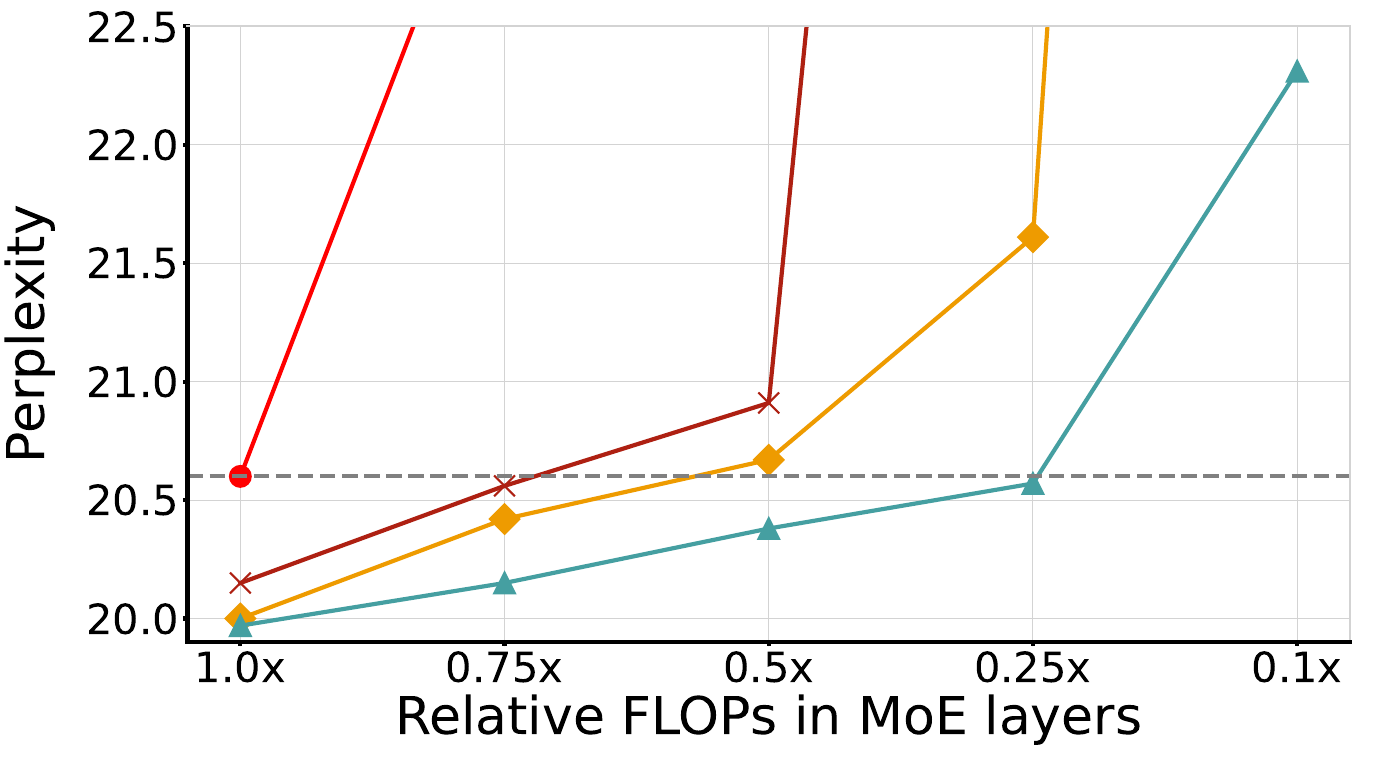}}
    \vspace{-0.5em}
    \caption{Test perplexity (PPL) with regard to the the normalized FLOPs in the MoE layer during inference. 
    We adjust FLOPs by modifying the capacity factor $\gamma$.
    }
    \label{fig:4_ppl_flops}
\end{figure}

\noindent \textbf{Efficiency.} 
Figure \ref{fig:4_ppl_flops} illustrates the impact of the number of floating point operations (FLOPs) within MoE layers on perplexity score. 
We see that models exhibit poorer performance as FLOPs decrease. This trend is expected, as limited expert capacity can lead to more dropped tokens, thereby increasing the number of inadequately processed tokens.

It is observed that \tool with small expert size is robust to the reductions in FLOPs.
Notably, \tool with an expert size of $384$ outperforms Switch Transformer (referred to as Exp$8$-Size$3072$) on WikiText-103, while consuming only $25\%$ of the FLOPs of the latter. This finding suggests that a significant portion of parameters in models with large expert sizes are unnecessarily engaged in computations. By replacing these large experts with smaller ones, \tool not only improves performance but also saves computational resources.

\begin{table}[h!]
\centering
\resizebox{\linewidth}{!}{
\begin{tabular}{lccccc}
\toprule[0.10em]
\textbf{Method} &  \textbf{\# Experts} &  \textbf{Size} &  \textbf{FLOPs}  & \textbf{OpenWeb} & \textbf{Wiki} \\
\midrule[0.08em]
Dense & $1$ & $3072$ & $1.00$x  &  $22.61$ & $22.24$\\
\midrule[0.08em]
 

\multirow{8}{*}{\tool} 
& \multirow{4}{*}{$8$} &  \multirow{4}{*}{$384$}
& $1.00$x  & $22.89$ & $21.93$\\
& & & $0.75$x  & $22.89$ &  $21.93$\\
& & & $0.50$x  & $22.90$ &  $21.93$\\
& & & $0.25$x  & $23.07$ &  $21.97$ \\

\cmidrule[0.08em](l){2-6} 

& \multirow{4}{*}{$16$} &  \multirow{4}{*}{$192$}
& $1.00$x  & $22.94$ & $21.92$\\
& & & $0.75$x  & $22.97$ & $21.92$\\
& & & $0.50$x  & $23.02$ & $21.92$\\
& & & $0.25$x  & $23.27$ & $21.93$\\

\bottomrule[0.10em]
\end{tabular}
}

\caption{Test perplexity on OpenWebText and WikiText-103.}
\label{tab:dense_ppl}
\end{table}

 






\noindent \textbf{Dense Model.} Dense models can be trained as \tool by partitioning their FFN layers into smaller ones. As shown in Table \ref{tab:dense_ppl}, these modified models exhibit a remarkable reduction of over $50\%$ in computational complexity with only a marginal decrease in performance across both datasets. On the WikiText-103 dataset, \tool can outperform the dense model while achieving a $75\%$ reduction in Floating Point Operations (FLOPs). These findings underscore the potential of sparse models as promising alternatives to their dense counterparts.

\begin{table*}
\centering
\resizebox{\linewidth}{!}{
\begin{tabular}{lcc|cc|cc|cc|cc|cc|cc}
\toprule[0.10em]

\multicolumn{3}{c}{} & \multicolumn{2}{c}{\textbf{Uk-En}} & \multicolumn{2}{c}{\textbf{De-En}} & \multicolumn{2}{c}{\textbf{Ru-En}} & \multicolumn{2}{c}{\textbf{He-En}} & \multicolumn{2}{c}{\textbf{Zh-En}} & \multicolumn{2}{c}{\textbf{Avg}} \\
 \textbf{Method} & \textbf{\#Experts} & \textbf{Size} & \textit{BLEU} & \textit{ChrF} & \textit{BLEU} & \textit{ChrF} & \textit{BLEU} & \textit{ChrF} & \textit{BLEU} & \textit{ChrF} & \textit{BLEU} & \textit{ChrF} & \textit{BLEU} & \textit{ChrF} \\
 
 
\midrule[0.08em]
Dense & $1$ & $2048$ & $29.2$  & $53.0$ &  $28.2$ & $52.2$ & $25.3$ & $50.9$ & $32.8$ & $58.0$ & $16.7$ & $43.4$ & $26.4$ & $51.5$\\
Switch & $32$ & $2048$ & $29.9$ & $53.9$ &  $29.2$ & $53.1$ & $26.7$ & $52.0$ & $34.9$ & $59.7$ & $17.4$ & $44.8$ & $27.6$ & $52.7$ \\
\midrule[0.08em]

\multirow{2}{*}{Top-$k$} & $64$ & $1024$ & $30.7$ & $54.6$ &  $29.2$ & $53.1$ & $27.4$ & $53.5$ & $35.9$ & $\mathbf{60.9}$ & $17.6$ & $44.9$ & $28.2$ & $53.4$ \\
 & $128$ & $512$ & $30.9$ &  $54.9$ & $29.3$  & $53.1$ &  $27.8$ & $53.5$  & $\mathbf{36.3}$  & $\mathbf{60.9}$ & $17.7$ & $45.1$ & $28.4$ & $53.5$\\
\midrule[0.08em]
\multirow{2}{*}{\tool} & $64$ & $1024$ & $31.7$ & $\mathbf{55.7}$ &  $\mathbf{29.4}$ & $\mathbf{53.3}$ & $28.4$ & $\mathbf{54.0}$ & $36.1$ & $60.5$ & $\mathbf{18.1}$ & $\mathbf{45.6}$ & $\mathbf{28.7}$ & $\mathbf{53.8}$\\
& 128 & 512 & $\mathbf{32.0}$ & $\mathbf{55.7}$ & $29.2$ & $53.2$ & $\mathbf{28.5}$ & $53.9$ & $35.5$ & $60.1$ & $18.0$ & $45.4$ & $\mathbf{28.7}$ & $53.7$\\




\bottomrule[0.10em]
\end{tabular}
}

\caption{Machine translation on WMT23 datasets. \textit{ChrF} is the abbreviation of ChrF$2^{++}$}.
\label{tab:trans}

\end{table*}







\subsection{Machine Translation}

We compare \tool with Switch Transformer and Top-$k$ routing. Models are trained to translate other languages to English. All models have the same number of shared parameters and have the same FLOPs. The results are detailed in Table \ref{tab:trans}, revealing that \tool outperforms its counterparts. The observations generally align with the finding on language modeling that a reduction in expert size can always lead to better performance. However, the extent of the improvement depends upon factors such as the specific language being translated, and the evaluation metrics employed.

\subsection{Analysis}
\begin{figure}[t]
    \centering
    \includegraphics[width=0.45\textwidth]{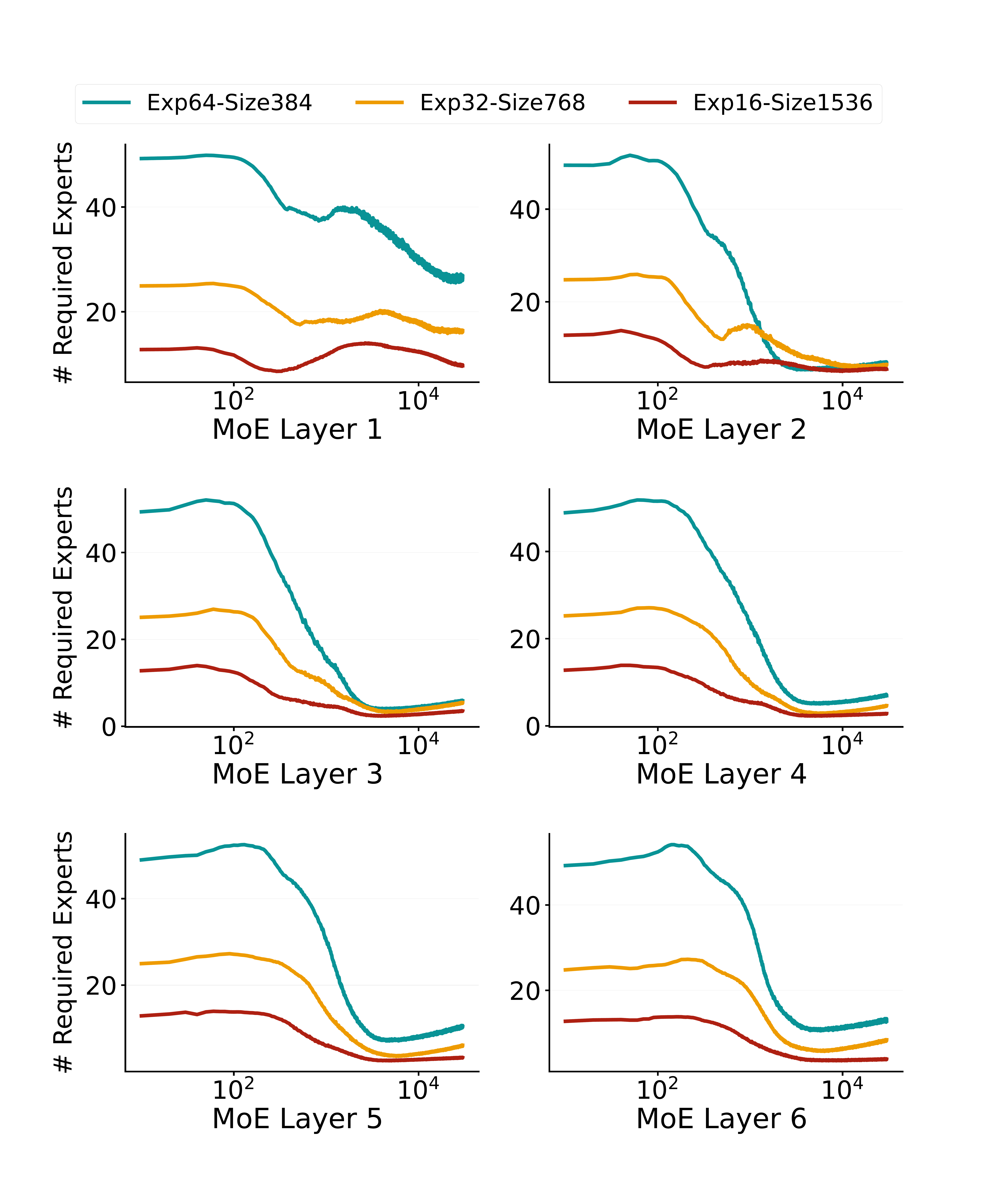}
    \caption{Average number of required experts with regards to training steps across different layers.}
    \label{fig:moe_want_num}
\end{figure}

\textbf{Efficiency.} \label{sec:router_ana}
Figure \ref{fig:moe_want_num} shows the number of experts that tokens require to satisfy the threshold vs. training step. Initially, a substantial portion of experts is required to satisfy the threshold, but as training progresses, a significant reduction in the required number of experts is observed. This phenomenon is consistent with existing research indicating that dense models exhibit sparse activation once trained \cite{li2023the}.
Our proposed \tool leverages this emergent sparsity to to enhance computational efficiency by employing small experts and a threshold-based router. The use of small-scale experts facilitates fine-grained expert selection, thereby mitigating the activation of redundant parameters. Concurrently, the threshold-based router encourages tokens to select the fewest experts. While the threshold-based routing may initially lead to an excessive selection of experts, the computations is supposed to exhibit sparsity after training.
\medskip

\begin{figure}[t]
    \centering
    \includegraphics[width=0.45\textwidth]{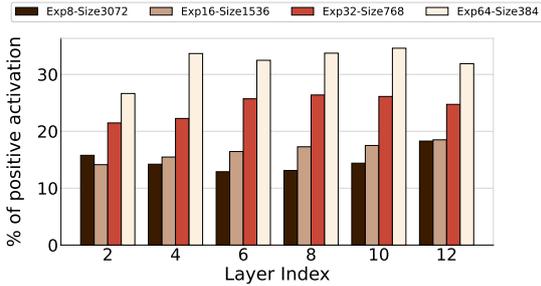}
    \caption{Average percentage of positive values in the FFN layers after the activation function.}
    \label{fig:eff_evidence}
\end{figure}

\noindent  \textbf{Effectiveness.}
The visualization in Figure \ref{fig:eff_evidence} illustrates the average percentage of positive values after the activation function across different configurations.
It can be seen that a reduction in expert size consistently correlates with an increase in this percentage.
This trend suggests that models with small experts can more effectively leverage the parameters within selected experts, thereby yielding performance improvements.

\subsection{Hyperparameters}

\begin{figure}[t]
    \centering
    \includegraphics[width=0.43\textwidth]{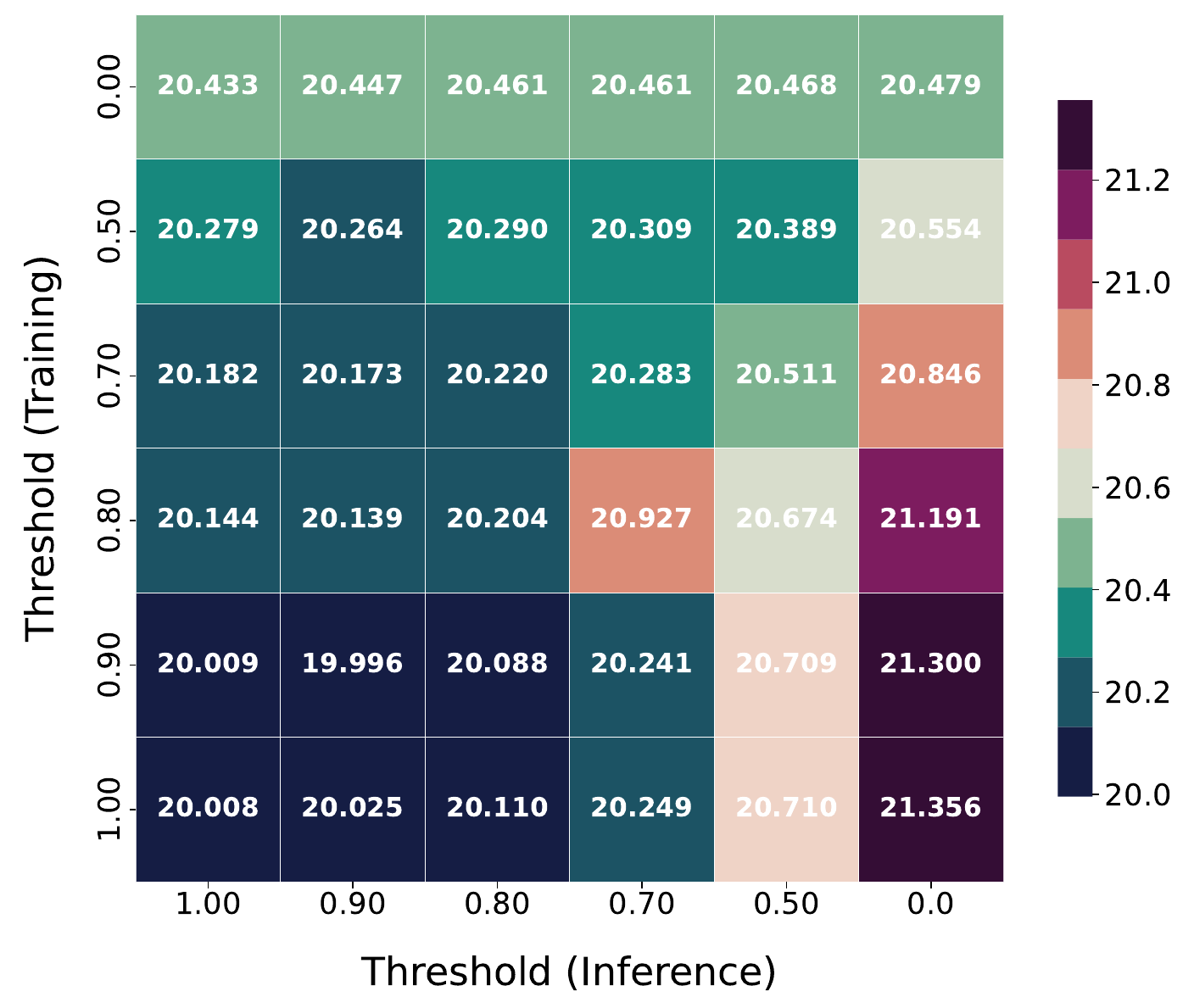}
    \caption{Effect of threshold on $\textrm{perplexity}^{\downarrow}$  during the training and inference stages. The models are trained on the WikiText-103 dataset utilizing 32 experts, each with a size of 768.}
    \label{fig:threshold_hyper}
    \vspace{-1.0em}
\end{figure}

\begin{figure}[h]
    \centering    \includegraphics[width=0.45\textwidth]{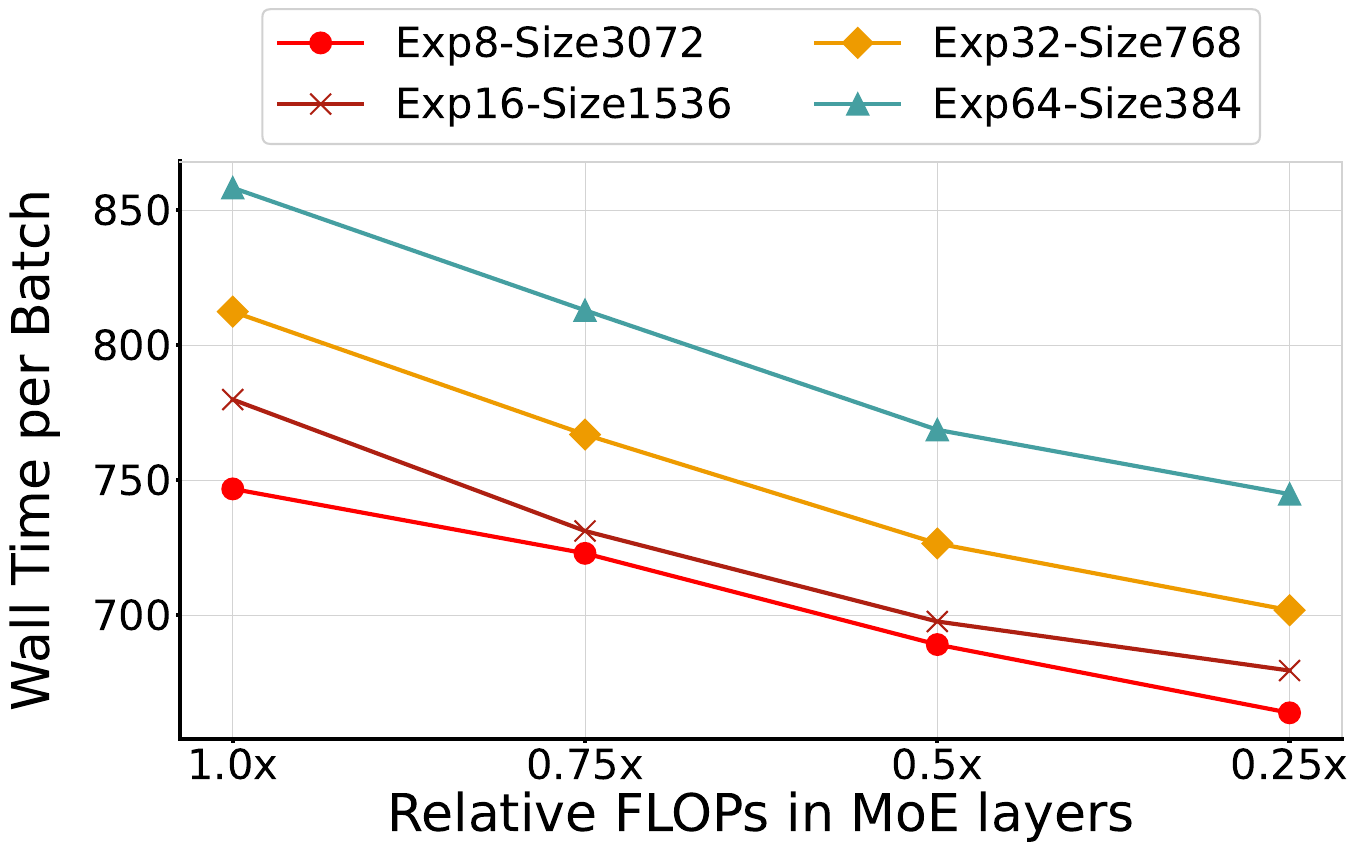}
    \caption{Per-batch wall time vs. FLOPs at MoE layers.}
    \label{fig:wall_time}
     \vspace{-1.0em}
\end{figure}

\noindent \textbf{Threshold.} 
We investigate the effect of threshold $t$ on the perplexity score during both training and inference stages. From Figure \ref{fig:threshold_hyper}, we see that increasing the threshold generally improves performance. However, \tool with a threshold of $0.9$ slightly outperforms the model with a threshold of $1.0$. 
It is noteworthy that at a threshold of $1.0$, each token is sent to all experts by the threshold-based router. In contrast, at a threshold of $0.0$, the router behaves like a top-$1$ router. 
Hence, replacing the top-$1$ router with the top-$N$ router, where $N$ represents the number of experts, is also a simple approach to enhance performance.
\medskip

\noindent \textbf{Wall Time.}
Figure \ref{fig:wall_time} illustrates the per-batch wall time during inference. We observe that  a reduction in FLOPs at the Mixture of Experts (MoE) layers correlates with a decrease in overall wall time. However, \tool with smaller experts exhibit significantly higher latency compared to those with larger experts. The reason is that smaller experts require more sparse computation, which is not well supported on computation hardware such as TPUs and GPUs \cite{li2023the}. In addition, the increased number of experts introduces additional computational overhead to the routing process due to operations like sorting and ranking across the expert pool. These observations highlight the limitations associated with further reducing expert dimensions.
The size of experts should be properly chosen in order to fully utilize the advantages offered by \tool.
\medskip

\section{Conclusion}
This paper proposes a novel MoE design, \tool, with the primary objective of improving the efficacy and efficiency of sparse MoE models. By employing small experts and a threshold-based router, \tool demonstrates performance enhancements while significantly reducing FLOPs, leveraging the inherent sparsity of the model. Our research sheds light on the utilization of sparsity to improve model quality. As for future directions, we aim to further harness the advantages of sparse computation, focusing on enhancements from both hardware and algorithmic perspectives.

\section{Limitations}
Our experiments were conducted on language modeling and machine translation tasks.
To ascertain the effectiveness of \tool across a broader spectrum of NLP tasks, additional experiments are necessary. 
Additionally, due to the limited computational resources in our experiments, the largest model explored in this paper comprises $556$ million parameters, notably smaller than the parameter counts in prevalent large-scale language models, which often exceed billions.  To substantiate the claims made in this paper, further investigations in larger-scale configurations are needed. The expert size is also an important factor to \tool, and setting it to $1$ could yield valuable insights. Regrettably, our current implementation makes this model unfeasible. We will leave these further investigation as our future work.

\bibliography{anthology,custom}

\appendix

\section{Appendix}
\label{sec:appendix}

\subsection{Expert Selection} \label{app:exp_sel}
The process for expert selection concerning a given token is presented in  Algorithm  \ref{alg:book_selection}. Tokens are dispatched to experts according to $I$. Subsequent to the routing phase, each expert independently handles the tokens assigned to them.
Given that the capacity of each expert is restricted to $C$, only the top $C$ tokens, as per the priority $R$, are processed by each expert. Any tokens beyond capacity are disregarded.

\begin{algorithm}[h]
\caption{Expert Selection Procedure}
\label{alg:book_selection}
\begin{algorithmic}[1]
\REQUIRE Probability distribution $p = [p_1, p_2, ..., p_n]$, Threshold $t$
\ENSURE Indices of selected experts $I$, corresponding priorities $R$
\STATE Sort $p$ in descending order
\STATE Initialize $I = []$, $R = []$ and $sum = 0$
\FOR{$i=1$ to $N$}
    \STATE $sum = sum + p_i$
    \STATE Append $i$ to $I$
    \STATE Append $p_i - i$ to $R$
    \IF{$sum \geq t$}
        \STATE Break
    \ENDIF
\ENDFOR
\RETURN $I$, $R$
\end{algorithmic}
\end{algorithm}

\subsection{Load Balancing Loss} \label{app:balance_loss}
Let $N$ represent the total number of experts involved in the evaluation process. The auxiliary loss function is formulated as follows:
\begin{align}
    \textrm{loss} = N \cdot \sum_{i=1}^{N} f_i \cdot p_i,
\end{align}
where $f_i$ denotes the fraction of tokens that rank the $i$-th expert as their top choice, and $p_i$ represents the sum of probabilities assigned to the top-ranked selection by these tokens.

\begin{table}
\centering
\begin{tabular}{lccccc}
\toprule[0.10em]
Code &  Language &  $\#$Bitext & Test \\
\midrule[0.08em]
Uk & Ukrainian & $10$M & flores$200$\\
De & German & $30$M & WMT$22$\\
Ru & Russian  & $10$M & WMT$22$\\
He & Hebrew  & $10$M & flores$200$\\
Zh & Chinese & $30$M & WMT$22$\\
\bottomrule[0.10em]
\end{tabular}
\caption{Statistics of the training resources X$\rightarrow$En from WMT23.}
\vspace{-1.0em}
\label{tab:trans_data}
\end{table}

\end{document}